\documentclass[jsa]{iosart2x}
\pdfoutput=1
\usepackage{xcolor}
\usepackage{nicefrac}
\usepackage{xspace}
\usepackage{graphicx}
\usepackage{amsmath}
\usepackage{mathtools}
\usepackage{amsfonts}
\usepackage[utf8]{inputenc}
\usepackage{enumitem}
\usepackage{nameref}

\usepackage[capitalize]{cleveref}
\usepackage{multirow}
\usepackage{float}
\usepackage{acronym}
\usepackage{epstopdf}
\usepackage{comment}
\usepackage{afterpage}
\usepackage{verbatim} 
\usepackage{hyperref}
\usepackage{natbib}

\usepackage[switch,pagewise]{lineno} 
\graphicspath{{./plots/}}
\newcommand{\figref}[1]{Fig.~\ref{fig:#1}}
\newcommand{\secref}[1]{Section~\ref{sec:#1}}

\iftrue

\newcommand{\todo}[1]{\textcolor{red}{\textbf{todo: #1}}}

\newcommand{\sophie}[1]{\textcolor{orange}{\textnormal{(Sophie) #1}}}

\else
\newcommand{\todo}[1]{}
\fi

\newcommand{\revision}[1]{#1}
\begin{document}

\begin{frontmatter}
\title{Supervised learning for table tennis match prediction}
\runtitle{Supervised learning for table tennis match prediction}

\begin{aug}

% removed for review
\author{\fnms{Yu-Hung Sophie} \snm{Chiang},\ead[label=e1]{schiangbc2821@gmail.com}
\thanks{Corresponding author: Sophie Chiang, \orgname{The Perse School, Hills Rd, Cambridge}, \cny{United Kingdom}, CB2 8QF. \printead{e1}.}}
\author{\fnms{Gyorgy} \snm{Denes}\ead[label=e2]{gdenes@perse.co.uk}
}

\address{\orgname{The Perse School, Hills Rd, Cambridge, United Kingdom, CB2 8QF}}

\end{aug}

\begin{abstract}
Machine learning, classification and prediction models have applications across a range of fields. Sport analytics is an increasingly popular application, but most existing work is focused on automated refereeing in mainstream sports and injury prevention. Research on other sports, such as table tennis, has only recently started gaining more traction. This paper proposes the use of machine learning to predict the outcome of table tennis single matches. We use player and match statistics as features and evaluate their relative importance in an ablation study. In terms of models, a number of popular models were explored. We found that 5-fold cross-validation and hyperparameter tuning was crucial to improve model performance. We investigated different feature aggregation strategies in our ablation study to demonstrate the robustness of the models. Different models performed comparably, with the accuracy of the results (61-70\%) matching state-of-the-art models in comparable sports, such as tennis. The results can serve as a baseline for future table tennis prediction models, and can feed back to prediction research in similar ball sports.
\end{abstract}

\begin{keyword}
\kwd{Machine learning}
\kwd{table tennis prediction}
\end{keyword}

\end{frontmatter}

%% \jvol{XX}
%% \jnum{XX}
%% \paper{1234567}
%% \pubyear{2020}
%% \publisheddate{xxxx 00, 0000}
%% \currentdate{xxxx 00, 0000}
%% \doiinfo{TQE.2020.Doi Number}

%\title{Supervised Learning for Table Tennis Match Prediction}
%\author{\IEEEauthorblockN{Yu-hung Sophie Chiang}, and
%    Gyorgy Denes\

%\thanks{Yu-hung Sophie Chiang and Gyorgy Denes are with the Perse School, Hills Rd, Cambridge CB2 8QF, UK (correspondence email: gdenes@perse.co.uk)}
%}

\if{false}  % Feel free to add here, but removing from manuscript
If not IEEE TAI, we could consider:
\begin{itemize}
    \item{Computers \& Mathematics with Applications (Elsevier)}
    \item{\sophie{MLSA 2022 : Machine Learning and Data Mining for Sports Analytics}}
\end{itemize}
\fi

% GXD: roughly the structure I was thinking, but feel free to tweak/move bits
\section{Introduction}
Table tennis is a quick and highly technical sport, requiring players to respond to an incoming ball trajectory within milliseconds. Rallies are intense, with ball speeds of 60--70mph and rotational speeds of 9000rpm. The proximity between players is  lower compared to similar sports such as tennis and badminton. The  outcome of a game can be influenced by subtle factors, which can be hard for a human to recognize.
%The possible inaccuracy in human decision making has thus lead to the application of machine learning techniques to sport result prediction.

Machine learning methods have been used frequently  in other sports, such as tennis and football. Proposed applications involve improved training efficiency 
%\denes{consider no citation, and repeat ref later in the text}
as well as result prediction. Specifically, result prediction is of high concern to sport fans, but little has been done in table tennis prediction.
%, the main reason being is the lack of available match data.  GXD: this is an important point, but here it almost implies that this paper adds dat

While manually-collected datasets alongside some analysis have been available in the past \citep{wang2019tac}, it is only recent developments in  multi-class event spotting and small object tracking that made accurate, detailed in-game data attainable.
In this paper, we propose using some of this freshly available data to train and evaluate  state-of-the-art classification algorithms on both men and women's professional singles matches. %Player statistics collected from historical matches are predominantly used in predicting the outcome, with newly derived information calculated from combining player statistics.

% moved to impact statement
%The main contributions of this paper are:
%\begin{itemize}
%    \item a quantitative evaluation of different ML models for table tennis match prediction
%    \item  an investigation on engineering new features from the raw data
%\end{itemize}

%The main contributions of this paper focuses on bridging the gap between data collection and predictive analysis in table tennis, discussing relevant state-of-the-art machine learning models and reporting the model with the highest predictive performance using rigorous quantitative evaluation. 

The paper is organized as follows: we first review relevant publications on sports prediction and describe the OSAI dataset on which our work is built. Then, we describe the proposed feature set. Finally, we evaluate and compare the performance of four different commonly used models and perform a feature ablation.
\section{Background}
%\subsection{Table tennis}
%\denes{not sure about section headers, but it would make sense to separate out the background from the ML. Also, can we have citations for the data, please?}
Table tennis is played by hundreds of millions of people world-wide, with almost 40 000 professionals registered with the International Table Tennis Federation (ITTF). Games are fast-paced and highly technical, but there are  factors that make it attractive for mathematical modelling: matches have only two possible outcomes (there are no draws).

A table tennis match consists of a sequence of sets; in a professional singles match, the first player to win best of seven sets wins the match. 
%In doubles (two teams of two players play against each other), the first team to win best of five sets wins the match. 
This paper will be looking at modelling professional singles matches only, thus there is no need to consider team line-ups.
In a set, the first player that earns at least eleven points and at least two more than their opponent wins the set. Each player serves twice before alternating, however, if the score reaches at least 10-10, each player serves only once before alternating.  
%The sport has proven to be very popular, with more than three hundred million players worldwide. %, a large proportion residing in East Asia.
The full set of rules are published by the \cite{ITTF}.

\section{Related Work} \label{sec:relatedwork}
\subsection{Machine Learning}
%\denes{We might want to just remove this subsection; it might be worth taking a look at some similar publications to check whether they even bother to include supervised learning in their related work (I suspect not)}.
Machine learning (ML) is a branch of artificial intelligence that has been successfully applied to many areas of industry and science, including disease diagnosis in medicine \citep{kourou2015machine}, pattern recognition \citep{weiss1989empirical}, computer vision \citep{khan2020machine} and bioinformatics \citep{larranaga2006machine}.
%This paper will be using supervised ML methods, as match instances are labelled and used to estimate the desired outcome.
The problem of predicting a table tennis match can be thought of as a supervised binary classification problem, with unambiguous ground-truth match outcome labels widely available. %as data is categorized into one of two possible classes.

\subsection{ML in Sports}
%\todo{A general overview of ML in sports}

In the past, manual data collection methods for sports have typically proven time-consuming and prone to human error and bias. Recent improvements in data capture have sparked interest in automatic data collection and analysis for a range of sports. \cite{xing2010multiple} proposed a dual-mode two-way Bayesian inference approach to track multiple highly dynamic and interactive players from videos in team sports such as basketball, football and hockey.
\cite{claudino2019current} used different ML methods, such as neural networks and decision tree classifiers, to investigate injury risk and performance in football, basketball, handball and volleyball.
\cite{davoodi2010horse} used neural networks for horse racing prediction, where eight features were used as input nodes to each neural network. This included information such as horse weight and race distance, to predict the eventual finishing time and rank of every horse in a race. 

% quite interesting, but potentially too much about horses
%Some horses lacked performance data, and had to be excluded from the analysis.

%therefore it would not be adequate to use it's history for prediction. As such, these horses had to be removed from the dataset to achieve better results.

Applications of ML in sports can help players and performance analysts in identifying critical factors that contribute to winning. Appropriate tactics can be identified in maximising player performance.
Aside from formulating strategies to win matches, using machine learning methods for sport result prediction has become popular due to the expanding domain in betting \citep{bunker2019machine}, which necessitates high predictive accuracy. Other applications include automated scouting and recruitment \citep{bunker2019machine} and umpiring assistance \citep{vzemgulys2018recognition}.

\subsection{Prediction in Tennis}
A number of data-driven models are available for \textit{tennis}. 
\cite{clarke2000using} predicted the outcome of professional tennis matches with 61--69\% accuracy using player rating points. Mapping player ability to a single rank can fail to capture complex factors, especially when comparing lower-rated players. In our paper we consider more complex features.

\cite{barnett2005combining} use rich historical data to predict the probability of a player winning a single point, building up a Markov chain to predict the winner of a match. The authors' approach is compelling, but there is no published data on the accuracy of their model on a larger dataset.

\cite{knottenbelt2012common} proposed a common opponent model to find a pre-play estimate of winning a match. This was achieved by analysing match statistics for opponents that both players encountered in the past. %, which provided for a fair basis comparison. 
The model computed the probability of each player winning a point on their serve, and hence the match. The authors found a 59--77\% accuracy, with an estimated return on investment of 6.85\% when put into the betting market for over four major tennis tournaments in 2011. 

%Due to similarities between tennis and table tennis, certain concepts can be applied from works that have been conducted in tennis.

%Both are ideal sports to apply hierarchical probability models to; a table tennis match consists of a sequence of sets, which consists of a sequence of points.

\subsection{ML in Table tennis}
In table tennis, ML research has focused so far on computer vision and automated data collection. % and  a number of computer vision based approaches have been applied. 
\cite{voeikov2020ttnet} proposed a neural network (TTNet) that allowed for real-time processing of high-resolution table tennis videos. They extract temporal and spatial data, such as ball detection and in-game events,  potentially replacing manual data collection by sport scouts. The model can also assist referees. More recently, \cite{zhang2010visual} used computer vision to allow a robot to play table tennis. They computed the 3D coordinates of a table tennis ball from a pair of video feeds to estimate the trajectory, the landing and striking point.

This paper builds on these existing works by utilising data by \cite{voeikov2020ttnet}, and applies it to the yet unexplored problem of supervised table tennis match prediction.
%, however, this study does not extend to predicting the match outcome.

%Current works in table tennis mainly focus on ball detection or calculating ball speed, where the focus has been on data collection rather than analysis. Predicting the result of a match has not yet been addressed, thus leading to the development of this project.

\section{Dataset} \label{sec:dataset}

\begin{figure}[ht]
\centering

\includegraphics[width=7.6cm]{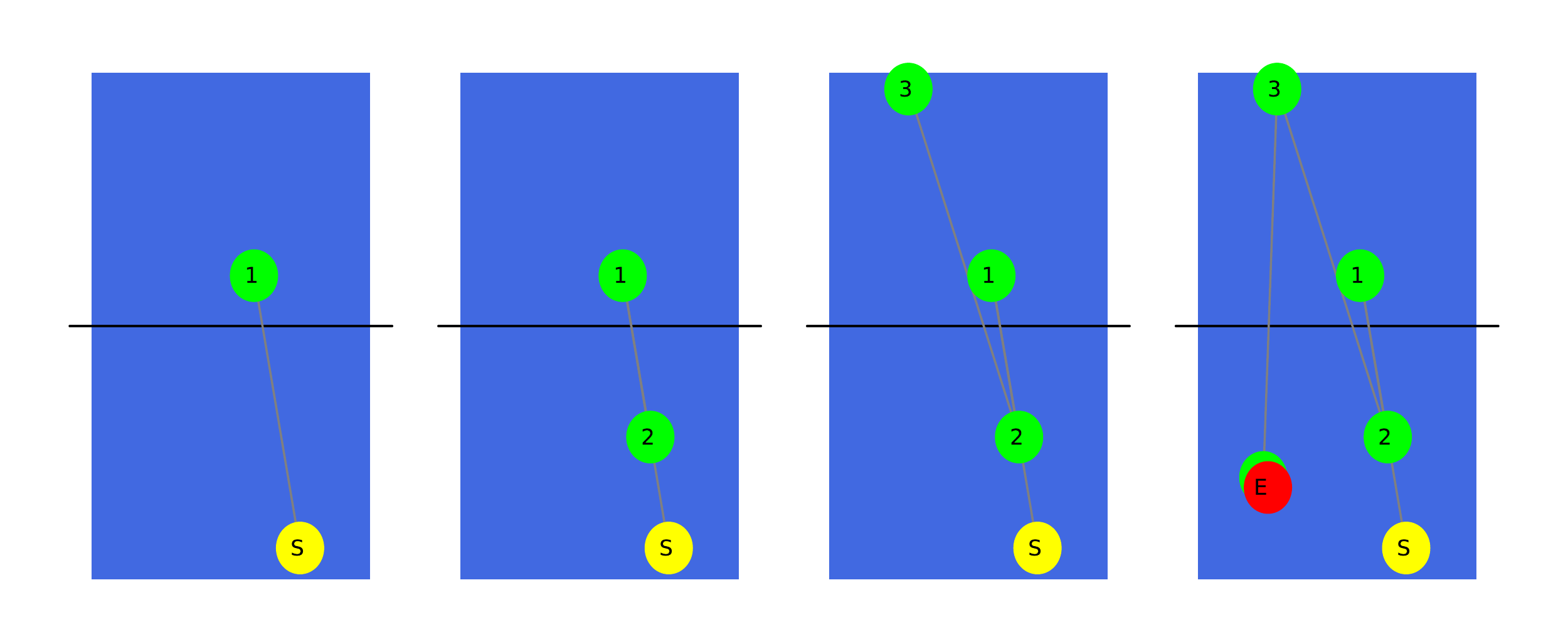}
\caption{Progression of a rally demonstrating the landing point of each ball bounce. Yellow indicates service which starts a rally and red indicates an error ending the rally. Green indicates all other ball bounces. Based on data from OSAI  \citep{OSAI}.}
%\denes{Please remove all margins on these plots}}

\label{fig:sequence}
\end{figure}

The primary source of data are automatic captures from TTNet \citep{voeikov2020ttnet}, released by \cite{OSAI}.
We use Tokyo 2020 Olympics and Tischtennis-Bundesliga (German table tennis league) data, which include men and women's singles matches.
Potential features include player  rank and in-match statistics such as percentage of points won on serve and receive, stroke  and error types. Match progression can be plotted for each set, recording the location of each ball bounce (see \figref{sequence}).

\begin{figure}[ht]
\centering

%\vspace{-2em}
\includegraphics[width=7.6cm]{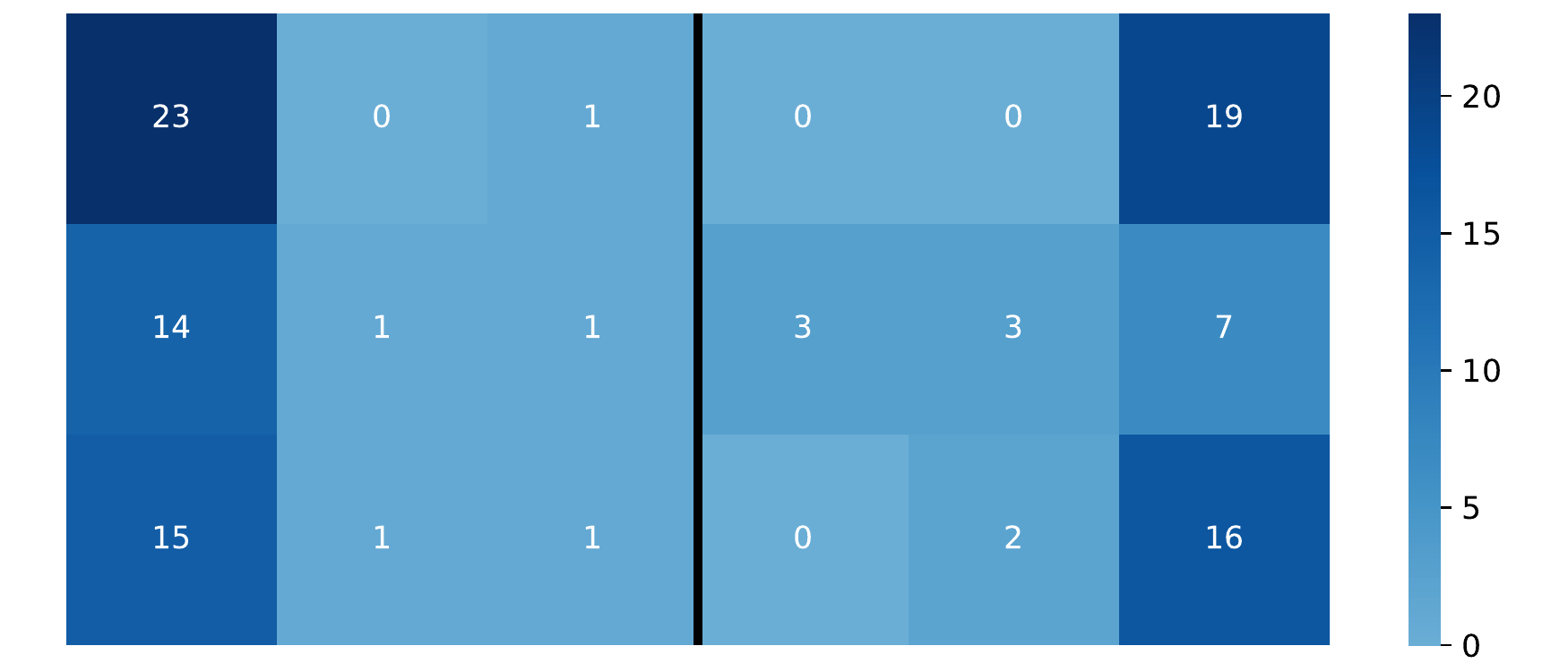}
\caption{Location of the last bounce of the winning ball, summed over an example match. Each side of the table is split into nine equal parts.}

\label{fig:pos}
\end{figure}

%Interactive maps that demonstrated the ball position of each shot on the table, as well as the stroke type were also accessible. The progression of a rally and location of each ball bounce can be mapped into a sequence (see \figref{sequence}).

\begin{figure}[ht]
\centering
\includegraphics[width=7.5cm]{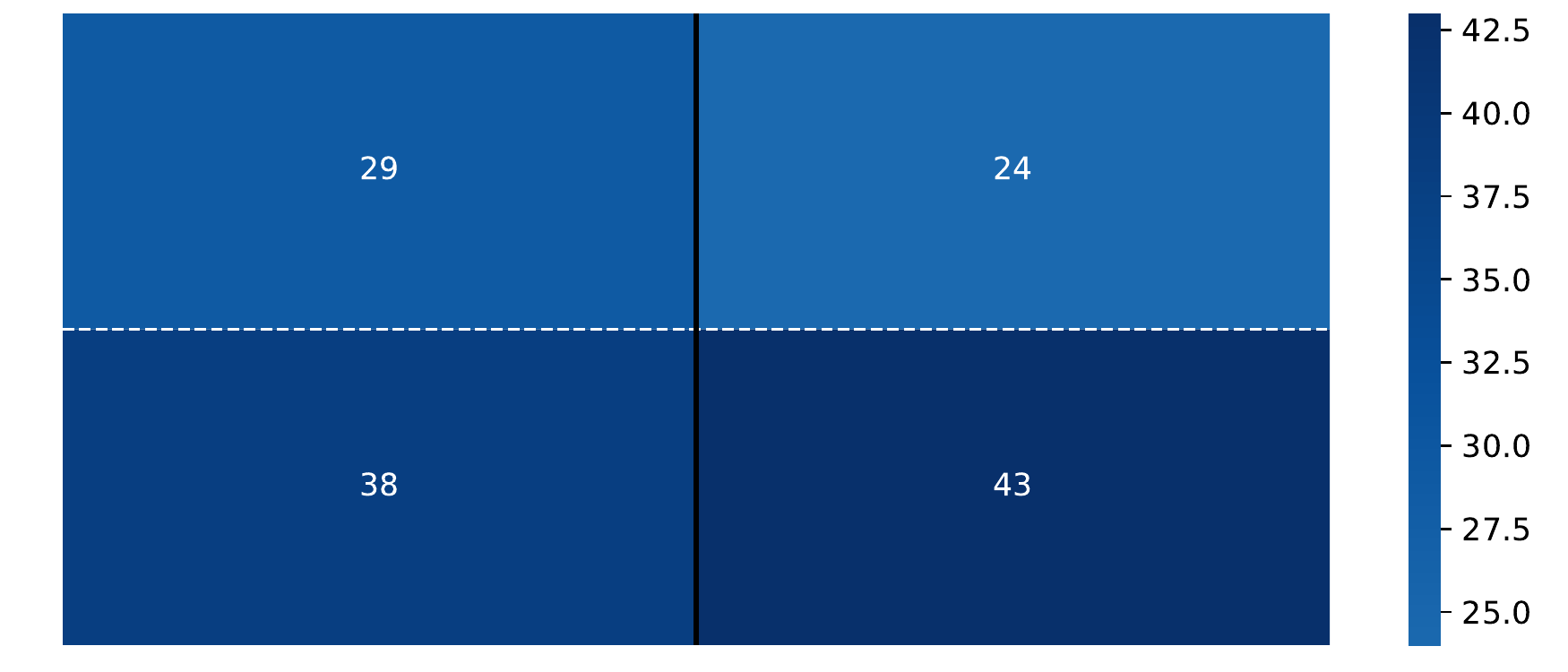}
\includegraphics[width=7.5cm]{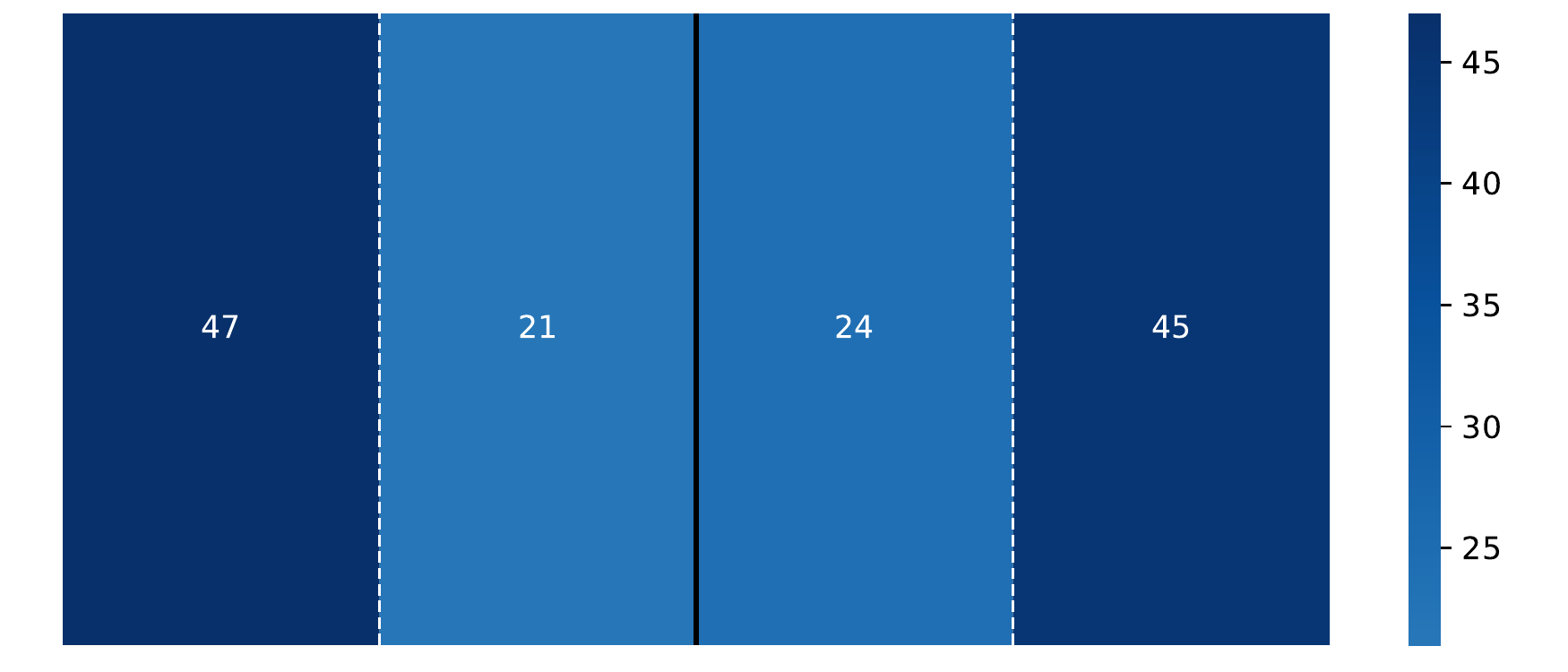}
\caption{Number of points won by forehand vs. backhand (top);  by a short vs. long rally (bottom) in an example match \citep{OSAI}.}

\label{fig:svlr}
\end{figure}

To reduce the dimensionality of the problem, a rally can be represented as the location of the winning shot. Furthermore, each half of the table can be split into nine equal sections, and the location of winning shots can be grouped (\figref{pos}). Further grouping can involve the number of forehands and the number of backhands used to win a point, or whether it was a `short' or `long' rally (\figref{svlr}). % The use of these features is discussed in  \secref{features}.
%\denes{which version? Fig 3 or 4?}.
Samples with missing data entries were removed from the dataset.

One of the main challenges in constructing a successful result predictor is the selection of salient features. To address this, features were carefully hand-picked and evaluated in \secref{experresults}.
%we carefully hand picked features that we thought would be the most influential in a match based on existing domain knowledge on the problem.
%\section{Features} \label{sec:features}
% Dr Denes: can keep, but removed for now to condense
%Rich data is valuable, however, irrelevant or redundant variables can increase both training and inference time, as well as decrease a model's performance, therefore choosing appropriate features to input into a classification model is very important.

\subsection{Match Representation}
We represent each match ($i$) from a participating player's ($P$) perspective as follows:
%In supervised machine learning, a set of labelled data is required for the model to train on. In the context of table tennis prediction, each match corresponded to two instances of data, one from the perspective of each player respectively, where every sample is composed of two elements:

\begin{itemize}
    \item a feature vector ($\underline{x}_i$) consisting of player and  match statistics,
    \item{
     the target variable ($y$), indicating the result of the match: %that corresponds to its respective sample
     \begin{equation}
        y_i =
        \begin{cases}
        1, &\text{if $P$ wins} \\ 
        -1, &\text{if $P$ loses}
        \end{cases}.
        \label{eq:y}
    \end{equation}
     }
\end{itemize}

With incomplete matches removed from the dataset, there are no other possible outcomes (there are no draws in table tennis). Each match actually maps to two ($\underline{x_i}$,$y_i$) pairs, from the perspective of the two participating players.
%As any incomplete matches were removed from the dataset, 

\subsection{Feature Engineering}
\label{sec:engineer}
An approach dictated by current state-of-the-art in \textit{tennis} was used
\citep{barnett2005combining,sipko2015machine,cornman2017machine}  to form new features for table tennis. These features are player-focused, as unlike in team sports, we do not need to consider line-ups, collective team ability or substitutions.

%Using pre-existing knowledge on the sport, adding combinations of player statistics as features may improve the predictive model. These features were calculated as differences between different player statistics as this considers the characteristics of \textit{both} players participating in a match. This was inspired by 

%In table tennis, there are only two possible outcomes of a match; a win or a loss. 

Table~\ref{table:nonlin} shows the final  set of features. Newly derived features are indicated with *. More detailed explanations can be found in the subsections below.
%\ref{sec:sp} to \ref{sec:balance}.

\begin{table}[ht]
\caption{Feature Summary} % title of Table
\centering % used for centering table
\setlength{\tabcolsep}{3pt}
\scalebox{1}{%
\begin{tabular}{c c} % centered columns (4 columns)
\hline\hline %inserts double horizontal lines
Feature & Explanation \\ [0.5ex] % inserts table

%heading
\hline % inserts single horizontal line
SP & percentage of total points won on serve \\
RP & percentage of total points won on receive \\
LRP & percentage of total points won on a long rally \\
SRP & percentage of total points won on a short rally \\
FHP & percentage of total points won on a forehand \\
BHP & percentage of total points won on a backhand \\ 
RANK & player ranking \\
RANKDIFF* & difference in rank between opponents \\
SA* & player serve advantage \\
SRA* & player short rally advantage \\
FHA* & player forehand advantage \\
BALANCE* & measure of how well rounded a player is \\
[1ex] % [1ex] adds vertical space

\hline %inserts single line
\end{tabular}}
%\vspace{4em}

\label{table:nonlin} % is used to refer this table in the text
\end{table}

\subsubsection*{SP: Serve Percentage}\label{sec:sp}
The proportion of points won on serve by a player. 
%This can be demonstrated from data similar to \figref{sequence}; 
If the serve and error are made on opposing sides of the table, it  was won by the serving player.

\subsubsection*{RP: Receive Percentage}\label{sec:rp}
The proportion of points won by the receiving player. E.g. in \figref{sequence}, the serve and error is made by the same player, so the point is won by the receiver.

\subsubsection*{LRP: Long Rally Percentage}\label{sec:lrp}
%\denes{ I'm confused again about the meaning of a long vs short rally}.
The proportion of points won on a long rally by a player to the total number of rallies won. In this paper we define a long rally as a rally of at least five shots. E.g. \figref{svlr} shows that one player won 47 points on a long rally, while the other won 45.
%\denes{how do we get percentage?}.

\subsubsection*{SRP: Short Rally Percentage}\label{sec:srp}
Compared to LRP, this is the proportion of points won on a \textit{short rally} by a player. E.g. \figref{svlr} shows that one player won 21 points on a short rally, while the other won 24 in the entire match.

\subsubsection*{FHP: Forehand Percentage}\label{sec:fp}
The proportion of points won on a forehand by a player, determined by the type of stroke used on the winning shot of a rally. \figref{svlr} shows one player won 24 points on a forehand, and the other winning 38.

\subsubsection*{BHP: Backhand Percentage}\label{sec:bp}
The proportion of points won on a backhand by a player. \figref{svlr} shows one player won 43 points on a backhand, and the other winning 29.

\subsubsection*{RANK} \label{sec:rank}
The ranking of the player by ITTF 
%\denes{who establishes the ranking? ITTF?}

\subsubsection*{RANKDIFF*: Rank Difference} \label{sec:rankdiff}
%Constructed by calculating the difference between rankings of two opponents:
\begin{equation}
    \text{RANKDIFF} = \begin{cases}
\text{RANK}_a - \text{RANK}_b &\text{for player $a$} \\
\text{RANK}_b - \text{RANK}_a &\text{for player $b$},
\end{cases}
\end{equation}
where RANK$_a$ and RANK$_b$ are  player rankings for payers $a$ and $b$ at the time of the match. A rank advantage (i.e. lower numerical value than an opponent's), yields a negative RANKDIFF. Rankings are mostly reliable for the top players; for example, players of rank 2 and 7 are more likely to have an accurate depiction of their relative ability than players of rank 150 and 155, despite the difference being identical. To account for this, we apply a simple non-linearity and set RANKDIFF to $0$ for matches where both players are ranked over 100.%, the feature  is ignored ($=0$). This is due to the fact that the lower the rank of a player, the more likely it is that there will be other players of a similar standard where rank doesn't accurately represent the standard of a player.

\subsubsection*{SA*, SRA*, HA*} \label{sec:advantage}
Serve advantage is calculated as the difference between their serve and receive winning percentage. This shows how likely a player is to win a point if they are serving rather than if they are on receive. Subsequently, the advantage a respective player has in a short rally over a long rally, as well as the advantage a respective player has in a forehand stroke over a backhand stroke, can be calculated.

\subsubsection*{BALANCE*} \label{sec:balance}

Players of a higher skill level tend to have fewer weaknesses and are stronger in more aspects of the game. We propose measuring the overall well-roundness as:

\begin{equation}
    \textit{BALANCE} = \frac{|\textit{SA}|+|\textit{SRA}|+|\textit{FHA}|}{3}.
\end{equation}

\subsection{Feature Scaling}
To account for the varying numerical range of the input features, and to make features comparable, we  each input is standardized to a zero mean with a unit standard deviation.
%Different features tend to have a varying range of values, therefore it is best practice to scale features as part of data pre-processing prior to learning. \textit{Standardization} is a scaling technique to centre values around the mean with a unit standard deviation \cite{bollegala2017dynamic}. We get a coded value by subtracting the mean of the sample from the data and dividing it by the standard deviation. The feature representing rank difference for example, is represented across a much larger range compared to features which are percentages.

\subsection{Live vs. aggregate data}

\begin{figure}[ht]
\centering
%\vspace{-2em}
\includegraphics[width=7.6cm]{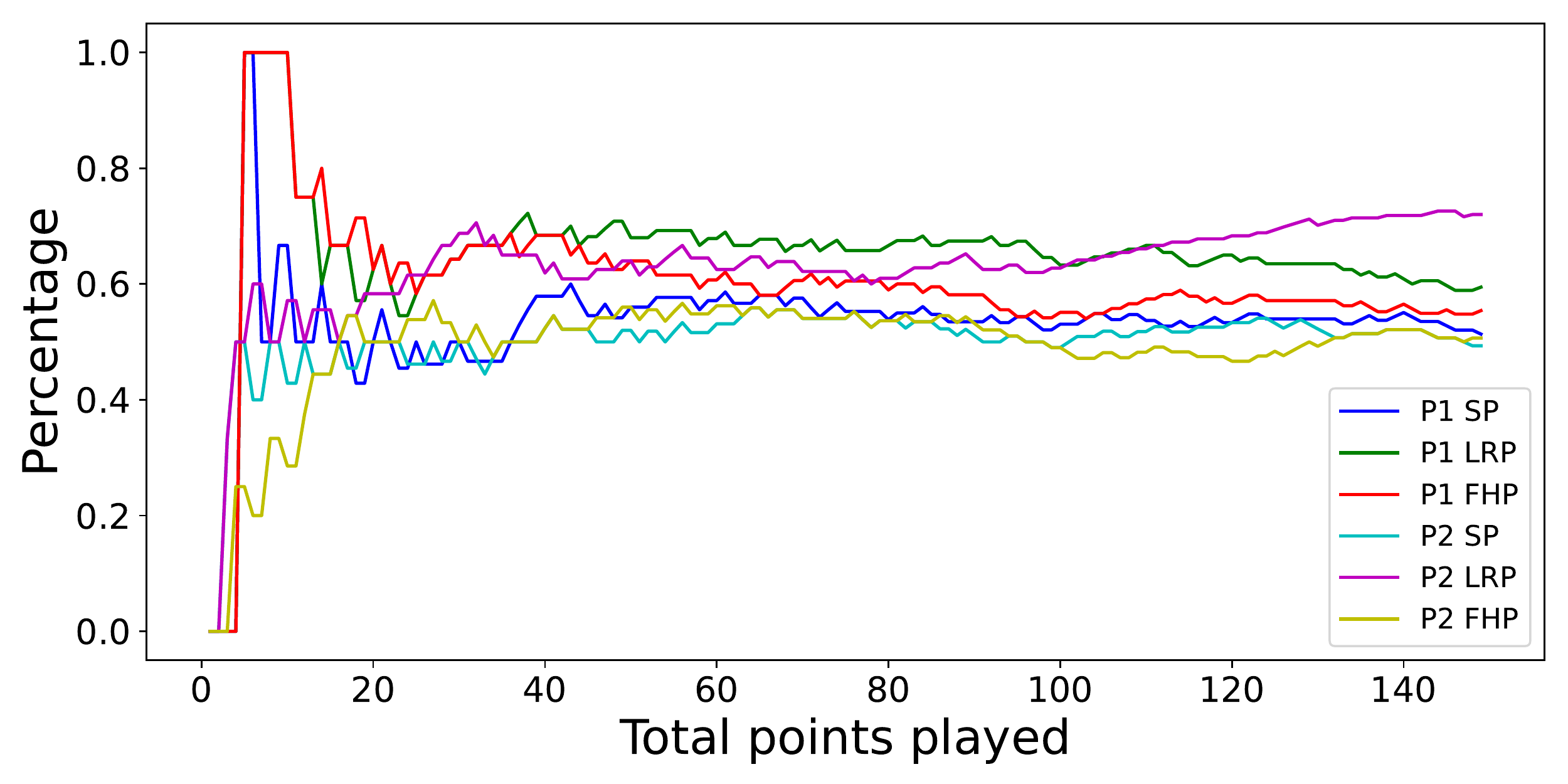}
\caption{Illustration of how individual features stabilize after a relatively short time on an example match. P1 and P2 correspond to the perspective of the two participating players.}
\label{fig:time}
\end{figure}

Each feature vector $\underline{x}_i$ represents a single match $i$ from a $P$ player's perspective with several features observed during that game (SP, RP, LRP, SRP, FHP, VHP, SA*, SRA*, FHA*, BALANCE*). \revision{Predicting the outcome from all these features post-match is a trivial task}; however, we postulate that the result of a match can be predicted from live in-match data well before the game ends, which is supported by the quick convergence of features as illustrated on an example match in \figref{time}. Hence, we fit our models on $(\underline{x}_i,y_i)$ pairs. We also validate the performance of these fitted models in Section~\ref{sec:ablation} with the feature vector $\underline{x}_i$ averaged over all past and future matches of player $p$ except for the target match. This approach is not robust for novice players who have very few data points and whose performance might change rapidly. We hope to address this limitation in future work.

\section{Models} \label{sec:models}

%\denes{This is where we can still cut out quite a lot if we want to}
Four ML models were evaluated (logistic regression, random forest, SVMs, MLPs) as implemented in  Scikit-Learn  \citep{pedregosa2011scikit}.

\subsection{Logistic Regression}
The logistic function maps the input feature $\underline{x}_i$ to a probability value $p_i$. Values over $0.5$ correspond to the player winning match $i$.  Training minimizes the \textit{logistic loss} function \citep{hazan2014logistic}:
\begin{equation}
    (p) = -\frac{1}{n} \sum_{i=1}^n p_i \log \left( \frac{y_i + 1}{2} \right) + (1-p_i)\log \left( \frac{1-y_i}{2} \right), 
\end{equation}
where $n$ is number of matches, $p_i$ is the predicted probability of a player winning match $i$, and  $y_i$ is as defined in Eq.~\ref{eq:y}.

%where a loss function measures the disparity between observations and their estimated fits .

\subsection{Random Forest}
Random forest classifiers consist of an ensemble of simpler decision trees $\{h(\underline{x},\theta_k),\ k=1,...\}$. %, where %the $\{\theta_k\}$ are independent identically distributed random vectors and 
%each tree casts a unit vote for the most popular class at input $\textit{x}$. 
For the $k$th tree, a random vector $\theta_k$ is generated and fitted to produce a classifier $h(\underline{x}, \theta_k)$ \citep{breiman2001random}. During inference, each tree  casts a vote from input $\underline{x}$; the output is decided by a majority vote. Decision trees tend to be simpler to interpret and quicker to train.

\subsection{Support Vector Machines (SVM)}
SVMs have been used for  \textit{tennis} match predictions.
These models identify the optimal hyperplane in the multi-dimensional feature space that separates data points into the two target classes (win, lose). During training, the marginal distance between this decision boundary and the instances closest to the boundary is maximized. %The existence of a decision boundary can allow for any detection of miss-classification. 
SVMs have a choice of \textit{kernels}, including linear, polynomial, sigmoid or a radial basis function
 \citep{cornman2017machine}.

\subsection{Multilayer Perceptron Neural Networks (MLP)}
An MLP is an artificial neural network consisting of an input layer ($\underline{x}$), an output layer (prediction), and  one or more hidden layers in-between. Neurons in consecutive layers are connected (no connections within layers) \citep{noriega2005multilayer}. Each connection has an associated weight. Training an MLP involves adjustments of these weights using backpropagation to minimize the difference between model output and the desired output.

\begin{figure*}[h]
\centering
\includegraphics[width=16cm]{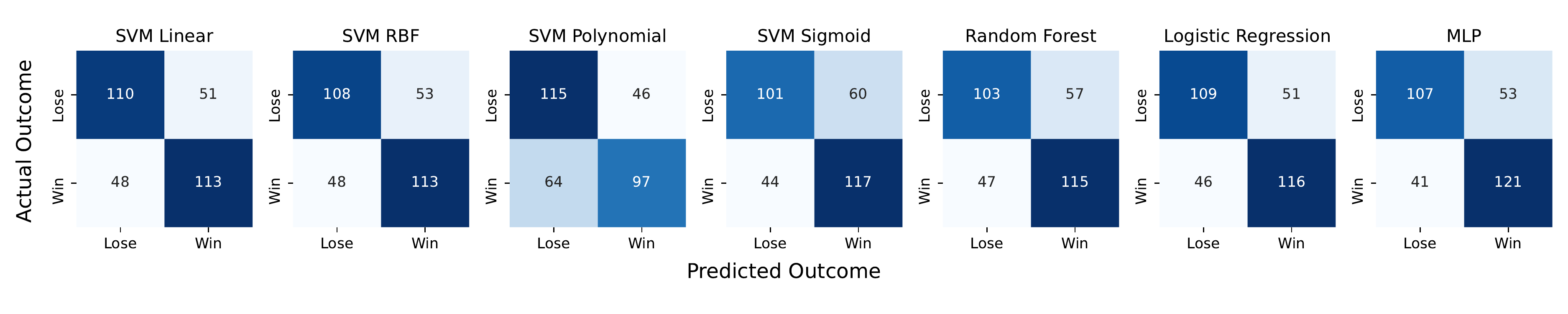}
\vspace{-1em}
\caption{Confusion matrices comparing the predicted and actual outcomes of test cases for each trained model.}
%\denes{Please reduce outside margins and consider removing the legend}}

\label{fig:confusionmatrices}
\vspace{-0.2cm}
\centering
\end{figure*}
\subsection{Evaluating Models} \label{evalmodels}
%\todo{shorten this}
To compare the performance of different model predictions, we calculated the \textit{accuracy} of each model
\begin{gather}
    \text{accuracy} = \frac{tp+tn}{tp+tn+fp+fn},
\end{gather}

where $tp$ and $tn$ are true positives and true negatives, and $fp$ and $fn$ are false positives and false negatives respectively.

%\sophie{should figures 2 and 3 be kept?}
%\denes{No, but we could report all the confusion matrices for each model (possibly in the appendices).}
To get a more balanced idea about model performance, we also compute F1 scores as:
%Precision is defined as the proportion of true positives to the total number of predictions predicted positive, and is a percentage of returned results which are relevant. Recall is defined to be the proportion of true positives to the total number of actual positives, and is a percentage of relevant data which have been correctly classified \citep{buckland1994relationship}.
%\begin{equation}
\begin{gather}
    \text{precision} = \frac{tp}{tp+fp} \quad
 \text{recall} = \frac{tp}{tp+fn}\\
 \vspace{0.0cm} \notag \\
F1 = \frac{2\times\text{precision} \times \text{recall}}{\text{precision} + \text{recall}},
\end{gather}
%\end{equation}
which is effectively an F measure with $\beta=1$ \citep{sokolova2006beyond}.
%Both metrics are important to take into consideration, and ultimately we use the F1 measure, which can be interpreted as the harmonic mean between precision and recall \citep{buckland1994relationship}, and accuracy score for evaluation of each classifier.
% An issue associated with training models is the possibility for the model to \textit{overfit}.7
During training, the model can over-fit to a specific dataset, which reduces model robustness. In cross validation, the dataset is split into $k$ random subsets, known as folds, and one is selected as a test set for the model to test on, while the others are used as a training set for the model to train on. This is repeated $k$ times where a different subset of data is used as the test set each time, and the overall performance of the model is calculated as the average of accuracy scores for each iteration \citep{berrar2019cross}. 
For model fitting here, 5-fold cross validation was used: the dataset was split in training:validation:test in a 72:18:10 ratio. 10\% of the original dataset was kept as a test set to validate hyperparameter tuning. The remaining 90\% of data was split in an 80:20 ratio for the 5-fold training; 80\% to train the model, 20\% to optimize hyperparameters. %The model trains on the training set, and the validation set is used for optimising the hyperparameters of the model.
%Overfitting occurs when a model captures unwanted bias and noise in the data that it negatively impacts the performance of a model. The model corresponds to it's initial training data too well, and fails to predict unseen data reliably. In order to limit overfitting, a popular re-sampling technique called $k$-fold cross validation is used.

\subsection{Hyperparameter Tuning} \label{hpt}
We used a brute-force grid search to fine tune parameters of the model that are outside the usual training domain (hyperparameters) e.g. the number of trees in a random forest classifier.
%For each model, the \textit{hyperparameters}, parameters that are not optimised by the training algorithm e.g. the number of trees in a random forest classifier, were tuned manually by using a grid search to test different values.
%A grid search uses brute force to search the entire space for different hyperparameter configurations.
The best combination of hyperparameters for a model is determined by whichever has the highest accuracy on the validation set using 5-fold cross validation.

\begin{figure}[ht]
\includegraphics[width=7.6cm]{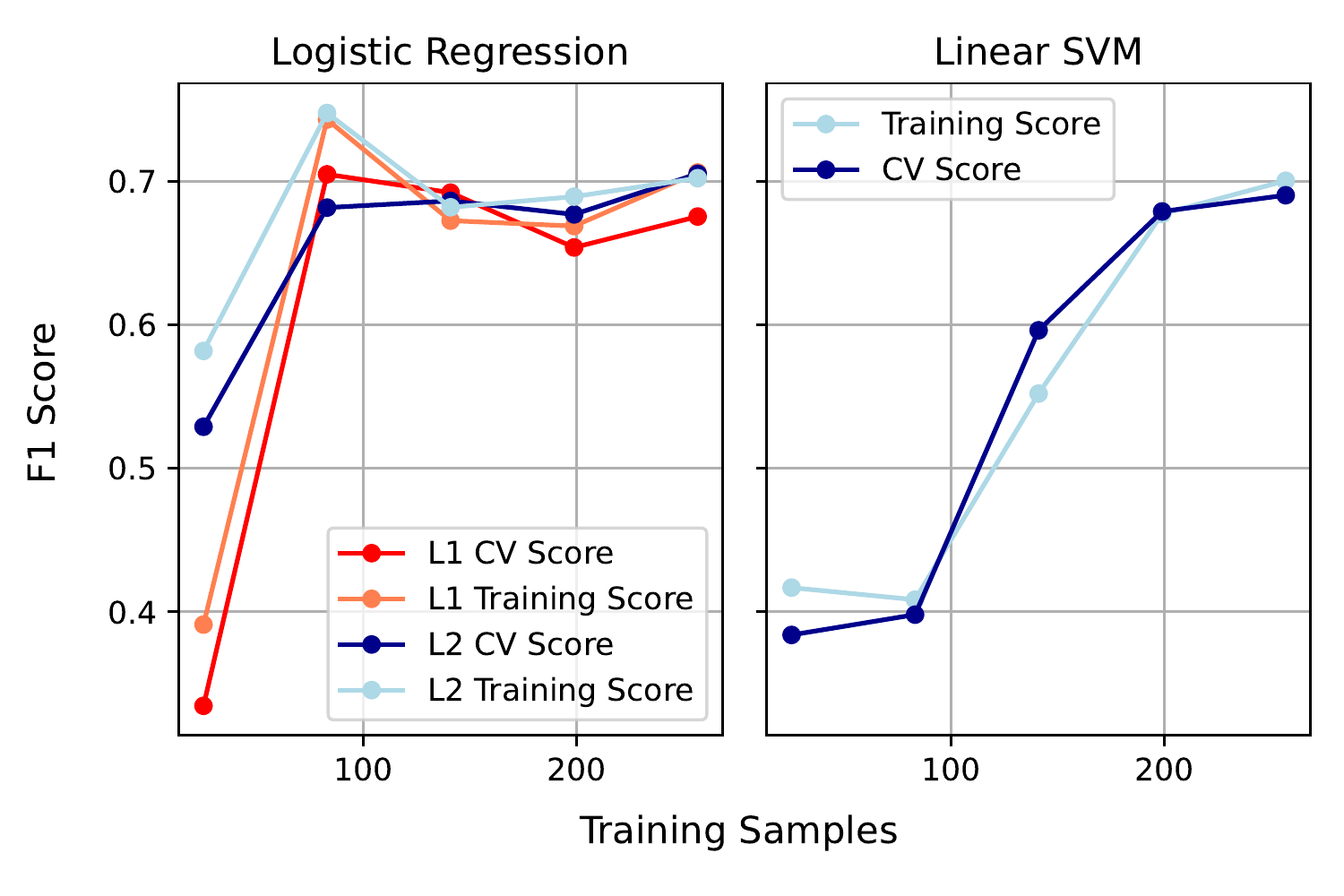}
\caption{Logistic regression and linear kernel SVM learning curves. The difference in F1 score for L1 and L2 regularisation is also illustrated.}
\label{fig:learningcurve}
\centering
\end{figure}

For logistic regression, the type of solver, penalty function and the $C$  terms value were adjusted. $C$ is a regularisation term; 
%\textit{Regularisation} prevents overfitting of training data by penalising large weights when training a logistic regression predictor, and the parameter $C$ mentioned is used to control the effect of this\citep{ahmadian1998regularisation}. 
the lower the value of $C$, the stronger the effect of regularisation. We found that  $C=1.0$, and `liblinear' solver resulted in the best average accuracy. In terms of regularisation, L2 regularisation gave better results than L1 (Fig. \ref{fig:learningcurve}).

%\denes{Could we plot l1 vs l2 here? That would be quite convincing}.

For SVMs, the two main hyperparameters that were adjusted were the kernel type and penalty value $C$. Using a linear kernel and $C=0.2$ gave the highest F1 score on the test set compared to other kernels. The learning curve for an SVM model using a linear kernel is shown in Fig. \ref{fig:learningcurve}.

\section{Experimental Results} \label{sec:experresults}
\begin{table}[ht]
\caption{\centering Model performance comparing validation and test sets before hyperparameter tuning}
\label{beforetuning}
\centering
\setlength{\tabcolsep}{3pt}
\scalebox{.9}{%
\begin{tabular}{ l|c c|c c }

\multirow{2}{4em}{Model} &
\multicolumn{2}{|c|}{Validation set} &
\multicolumn{2}{|c}{Test set} \\
\cline{2-5}
  & Acc & F1 & Acc & F1 \\

\hline \hline 
Logistic Regression & 0.696$\pm{0.025}$ & 0.701$\pm{0.025}$ & 0.694 & 0.645 \\
Random Forest & 0.658$\pm{0.018}$ & 0.658$\pm{0.028}$ & 0.639 & 0.667 \\
Support Vector Machine & & \\
$\rightarrow$ Linear & 0.680$\pm{0.019}$ & 0.679$\pm{0.030}$ & 0.611 & 0.611 \\
$\rightarrow$ RBF    & 0.680$\pm{0.022}$ & 0.681$\pm{0.029}$ & 0.611 & 0.611 \\
$\rightarrow$ Polynomial & 0.677$\pm{0.009}$ & 0.669$\pm{0.017}$ & 0.556 & 0.563\\
$\rightarrow$ Sigmoid & 0.590$\pm{0.022}$ & 0.553$\pm{0.028}$ & 0.694 & 0.579 \\
MLP Neural Network & 0.686$\pm{0.013}$ & 0.707$\pm{0.016}$ & 0.583 & 0.634\\
[1ex]
\hline
\end{tabular}}
\end{table}
\begin{table}[ht]
\caption{\centering Model performance comparing validation and test sets after hyperparameter tuning}
\label{results}
\centering
\setlength{\tabcolsep}{3pt}
\scalebox{.9}{%
\begin{tabular}{ l|c c|c c }

\multirow{2}{4em}{Model} &
\multicolumn{2}{|c|}{Validation set} &
\multicolumn{2}{|c}{Test set} \\
\cline{2-5}
  & Acc & F1 & Acc & F1 \\

\hline \hline 
Logistic Regression & 0.699$\pm{0.024}$ & 0.705$\pm{0.023}$ & 0.722 & 0.706 \\
Random Forest & 0.677$\pm{0.032}$ & 0.688$\pm{0.033}$ & 0.667 & 0.684 \\
Support Vector Machine & & \\
$\rightarrow$ Linear & 0.696$\pm{0.029}$ & 0.690$\pm{0.035}$ & 0.639 & 0.629 \\
$\rightarrow$ RBF    & 0.700$\pm{0.025}$ & 0.677$\pm{0.034}$ & 0.667 & 0.600 \\
$\rightarrow$ Polynomial & 0.705$\pm{0.021}$ & 0.685$\pm{0.021}$ & 0.611 & 0.563\\
$\rightarrow$ Sigmoid & 0.705$\pm{0.017}$ & 0.690$\pm{0.019}$ & 0.694 & 0.621 \\
MLP Neural Network & 0.696$\pm{0.019}$ & 0.708$\pm{0.020}$ & 0.694 & 0.703\\
[1ex]
\hline
\end{tabular}}
\end{table}

The main results are reported in Table~\ref{results} and \figref{confusionmatrices}.
%The test set is completely left out from the training process of the model, and is only used during evaluation. This is to see how well models generalise to unseen data which replicate new match data, as the test set is never used before evaluation. During evaluation, all other data is used for training the model (training and validation).
Both accuracy and F1 score are reported for the validation and test sets. The standard error for each score for the validation set is reported as a basis of defining uncertainty. The validation column shows that most models perform comparably with approx. 70\% accuracy. This value is also comparable to state-of-the-art metrics in \textit{tennis} match prediction. Table \ref{beforetuning} also indicates results across models before hyperparameter tuning (parameters will be set to default values), so that the difference can be compared.

F1 scores indicate that MLP Neural Networks (with a \textit{relu} activation) slightly over-perform their competitors, but the difference is not significant. The hidden layer size was set to 2 and the maximum number of iterations the solver iterates was chosen to be 200. The solver for weight optimization is set to `lbfgs', a quasi-Newton optimizer. The learning rate for scheduling weight updates is set to constant. However, the generic layered structure of a neural network has proven to be time consuming. Additionally, this technique is considered a `black box' technology, and finding out why a neural network has outstanding or even poor performance is difficult \citep{noriega2005multilayer}.

%On the test set, logistic regression seems to perform  noticeably better than other models. One interesting observation is that this is the only classifier that noticeably benefited the most from hyperparameter tuning. This might be partially due to the fact that we report the different kernels of SVMs separately.

\begin{figure}[ht]
\centering
\includegraphics[width=7.6cm]{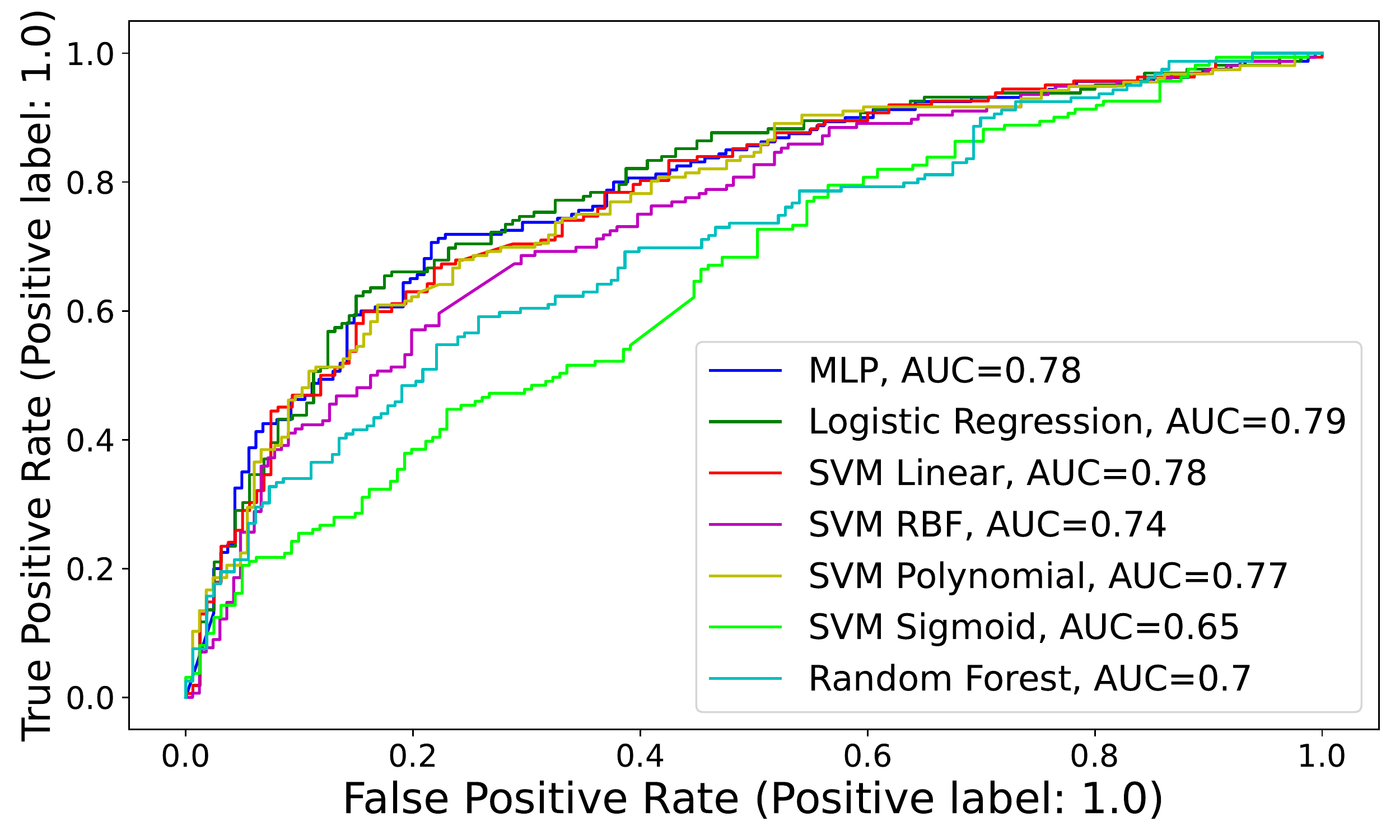}
\caption{ROC Learning Curves for the overall performance of each model.}
%\denes{As discussed, please replot without axis / bigger fonts. Or actually, why not do one single ROC plot (you have 4 colours anyway?}}

\label{fig:roc}
\end{figure}

Receiving operating characteristics (ROC, \figref{roc}) support the quantitative results. The kernel choice for SVM models makes a noticeable difference; the areas under the ROC curves are otherwise comparable for all other models.

%ROC curves demonstrate the performance of a classification model for all classification thresholds. The closer the apex of the curve is to the upper left hand corner, the greater the model's discriminatory ability \citep{fan2006understanding}. This is plotted for each type of classifier, including the SVM model using a linear kernel. 
%For each trained model, we use it's accuracy score, F1 measure and the area under its ROC curve to evaluate overall performance\denes{do we report area as well anywhere?}.

%\section{Discussion} \label{discussion}
%The model that achieved the highest F1 score on the validation set were MLP neural networks. The difference in accuracy and F1 score between the validation and test set was shown to be smaller in comparison to SVMs and random forest.

One qualitative advantage of using a random forest classifier is its training speed, which made hyperparameter tuning easier. The maximum number of levels in each decision tree was set to 80, the maximum number of features considered for splitting a node was set to 4, the minimum number of data points allowed in a leaf node was set to 4 and the number of trees that were in the classifier was set to 200.

\begin{figure}[ht]

\includegraphics[width=7.6cm]{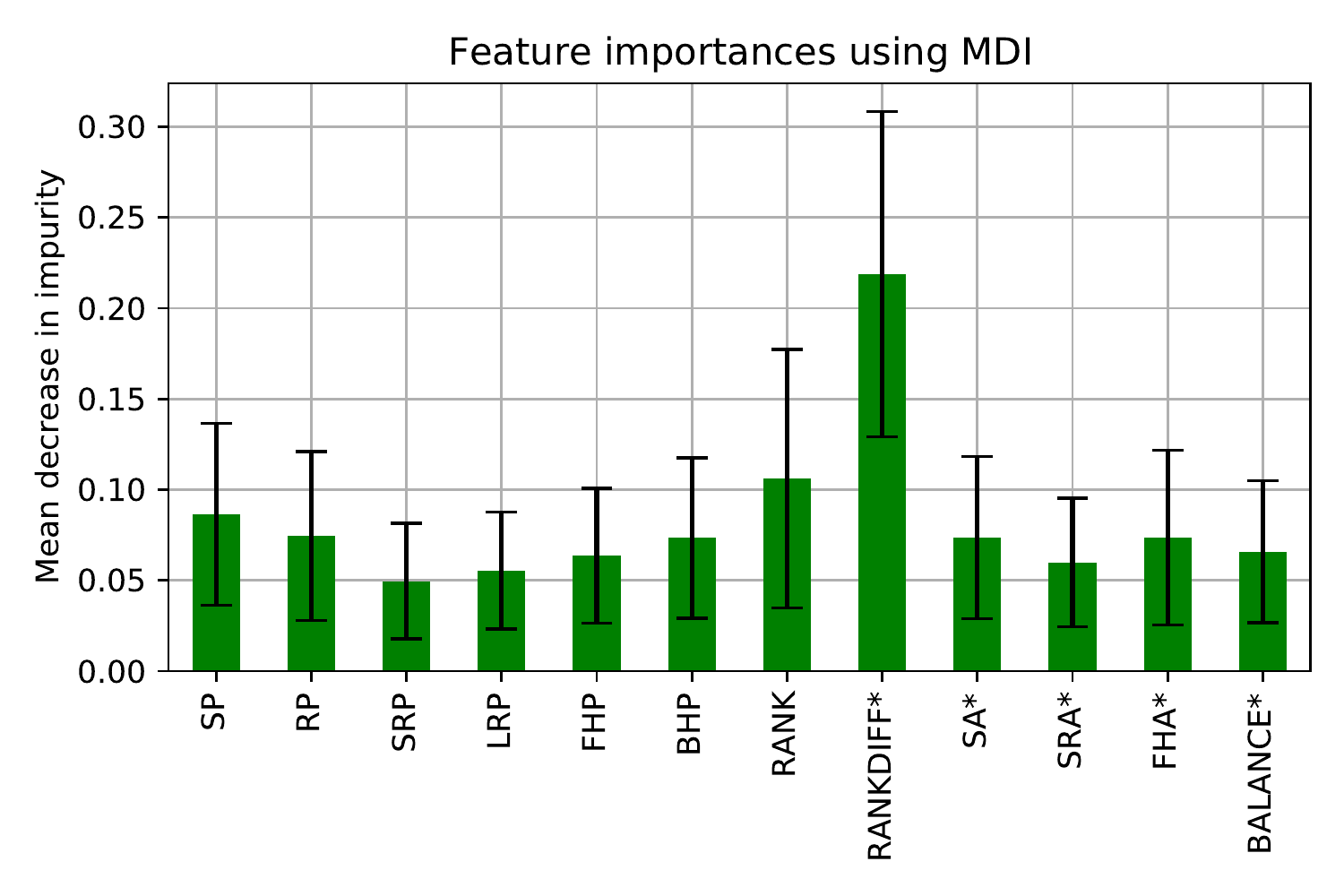}
\caption{Importance of features from random forest classifier based on Gini impurity. In our dataset, RANKDIFF appears to be the most important feature.}

\label{fig:fig4}
\centering
\end{figure}

Another significant advantage of a random forest classifier is that the importance of features can also be extracted and visualized.  \figref{fig4} shows this as the mean decrease in Gini impurity for features across all trees. The impurity of a node is the probability of a specific feature being classified incorrectly assuming that it is selected randomly \citep{cassidy2014calculating}.

\subsection{Ablation Study}
\label{sec:ablation}

\figref{fig4} predicts that the most important feature in a random forest model is  RANKDIFF, which justifies the inclusion of hand-crafted features. To reinforce this finding, we computed the accuracy and F1 score for each model with and without the derived features. All scores are lower for models that do not use newly derived features, and the accuracy score is significantly lower in SVMs compared to other models. See Table~\ref{results2}.

\begin{table}[ht]
\caption{Model performance with and without newly derived features}
\label{results2}
\centering
\setlength{\tabcolsep}{8pt}
\scalebox{0.9}{%
\begin{tabular}{ l|c c|c c }

\multirow{2}{4em}{Model} &
\multicolumn{2}{|c|}{With} &
\multicolumn{2}{|c}{Without} \\
\cline{2-5}
  & Acc & F1 & Acc & F1 \\

\hline \hline 
Logistic Regression & 0.699 & 0.705 & 0.631 & 0.668 \\
Random Forest & 0.677 & 0.688 & 0.661 & 0.673 \\
Support Vector Machine & & \\
$\rightarrow$ Linear & 0.696 & 0.690 & 0.556 & 0.619 \\
$\rightarrow$ RBF    & 0.700 & 0.677 & 0.500 & 0.591 \\
$\rightarrow$ Polynomial & 0.705 & 0.685 & 0.500 & 0.640\\
$\rightarrow$ Sigmoid & 0.705 & 0.690 & 0.472 & 0.642 \\
MLP Neural Network & 0.696 & 0.708 & 0.639 & 0.683\\
[1ex]
\hline
\end{tabular}}
\end{table}
\revision{A more real-world benchmark of performance is to understand how well the model can predict the outcome of a match before, without having seen that match}. We computed accuracy and F1 scores \revision{for this scenario} \revision{by restricting the model input to} aggregate feature vectors $\underline{x_i'}$ which contain the average features from all past and future matches, but exclude the target match $x_i$ itself. Table~\ref{prematch} shows similar results to live prediction, with accuracy values of 61--67\%. This indicates that our model is unlikely to have overfitted and just learnt existing match results and that it could be used as a robust match predictor.
\begin{table}[ht]
\caption{Model performance with the target match removed from the aggregate input features}
\label{prematch}
\centering
\setlength{\tabcolsep}{10pt}
\scalebox{1.1}{%
\begin{tabular}{ l|c c }

\multirow{2}{4em}{Model} &
\multicolumn{2}{|c}{Test set} \\
\cline{2-3} & Acc & F1 \\

\hline \hline 
Logistic Regression & 0.639 & 0.667 \\
Random Forest & 0.667 & 0.714 \\
Support Vector Machine & & \\
$\rightarrow$ Linear & 0.639 & 0.667 \\
$\rightarrow$ RBF    & 0.639 & 0.649 \\
$\rightarrow$ Polynomial & 0.611 & 0.632\\
$\rightarrow$ Sigmoid & 0.639 & 0.649 \\
MLP Neural Network & 0.667 & 0.714\\
[1ex]
\hline
\end{tabular}}
\end{table}
\section{Conclusion and Future Work} \label{sec:conc}
%Machine learning has proved suitable for solving numerous previously impossible tasks. 
This paper explores how supervised classification models can be used to predict the results of table tennis matches.  The original dataset was retrieved from OSAI \citep{OSAI}.

This paper utilises existing models as implemented in Scikit-learn (logistic regression, random forest classification, SVMs, multi-layer perceptrons); our contribution lies in applying these using 5-fold cross-validation and hyperparameter tuning to the problem of table tennis match prediction.
We also propose using a handful of engineered features, from which a non-linear rank difference has been proved to be the most salient in our ablation study. To investigate overfitting, We consider aggregating feature across all matches of a player including or excluding the target match and demonstrate that our model performs comparably in both cases. T

 Our results are comparable to the accuracy of state-of-the-art \textit{tennis} prediction models (approx. 70\% accuracy). Following hyperparameter tuning, the difference between models was often modest. Other considerations when picking a model for similar applications could include training time or model transparency (at both of which random forests excel). 
%Future works could focus on a selection of the most important features established from the random forest model.

Future work could explore combining TTNet with our prediction model to provide live match predictions. It would be also interesting quantifying uncertainty and to test against real betting odds. As automated table tennis analytics are becoming available below professional leagues, the authors are also interested whether the importance of features and the model choice transfers to these matches as well.

\section*{Acknowledgements}
The authors would also like to thank the OSAI team for granting permission to use their dataset.

\bibliographystyle{agsm}

\bibliography{References}
\end{document}